\documentclass[a4paper]{cas-dc}

\usepackage[numbers]{natbib}
\usepackage{graphicx}
\usepackage{float}
\usepackage{placeins}
\usepackage{caption}

\def\tsc#1{\csdef{#1}{\textsc{\lowercase{#1}}\xspace}}
\tsc{WGM}
\tsc{QE}
\tsc{EP}
\tsc{PMS}
\tsc{BEC}
\tsc{DE}

\begin{document}
\let\WriteBookmarks\relax
\def\floatpagepagefraction{1}
\def\textpagefraction{.001}
\let\printorcid\relax
\shorttitle{LG-PF: Lightweight Confidence-Guided Polarization Image Fusion}

\title [mode = title]{LG-PF: Lightweight Confidence-Guided Polarization Image Fusion}     

 \author[author1]{Zhuangfan~Huang}
 \ead{2112455033@stu.fosu.edu.cn}

 \author[author2]{Zhenyu~Kuang}
 \ead{kmustkzy@126.com}

 \author[author3]{Gao~Wang}
 \ead{wanggao@nuc.edu.cn}

 \author[author1]{Yang~Liu}
 \ead{ly25@fosu.edu.cn}
 \cormark[1]

 \author[author1]{Haishu~Tan}
 \ead{tanhaishu@fosu.edu.cn}

 \author[author1]{Xiaosong~Li}
\ead{lixiaosong@buaa.edu.cn}
 \cormark[1]

 \cortext[cor1]{Corresponding authors.}

 \address[author1]{Guangdong-HongKong-Macao Joint Laboratory for Intelligent Micro-Nano Optoelectronic Technology, School of Physics and Optoelectronic Engineering, Foshan University, Foshan 528225, China}

 \address[author2]{School of Electronic and Information Engineering, Foshan University, Foshan 528225, China.}

\address[author3]{State Key Laboratory of Dynamic Measurement Technology, North University of China, Taiyuan 030051, China}

\begin{abstract}
Polarization image fusion combines the stable luminance and structural information of the total-intensity image $S_0$ with the material-sensitive details of the degree of linear polarization ($DoLP$) image. However, the reliability of $DoLP$ varies spatially, and indiscriminate polarization transfer may amplify unstable responses or disturb the structural appearance anchored by $S_0$. We therefore propose LG-PF, a lightweight confidence-guided framework that formulates polarization fusion as a selective residual transfer process. A Polarization Confidence Prior estimates spatially reliable polarization responses, a Mask-guided Multi-scale Fusion module regulates their transfer across three feature scales, and a Lightweight Context-aware Bounded Correction Head stabilizes local photometric and structural transitions. Confidence guidance is also incorporated into the optimization objectives to preserve reliable polarization details while suppressing unsupported responses. We also construct MSP, a multi-scene polarization fusion dataset containing 1000 pixel-aligned image pairs from 17 indoor and outdoor scene categories. LG-PF achieves the best results across all six evaluated metrics on MSP, while subset-based evaluations on PIF and GAND show promising transferability without fine-tuning. With only $0.2936$ M parameters and an inference time of $21.712$ ms per image, LG-PF achieves competitive fusion quality with low computational cost.The source code will be available at https://github.com/1hzf/LG-PF.
\end{abstract}


\maketitle
\section{Introduction}
\label{sec:introduction}

Polarization imaging captures material- and reflection-dependent cues that are unavailable to conventional intensity measurements~\cite{yang2024polarimetricreview}. In the Stokes-based linear-polarization representation~\cite{collett1984stokes}, the total-intensity image $S_0$ provides a relatively stable luminance and structural reference, whereas the degree of linear polarization ($DoLP$) emphasizes complementary material responses, reflection boundaries, and fine-scale textures. However, the informativeness of $DoLP$ is spatially nonuniform, and its responses may become less reliable in low-intensity, weakly polarized, or noise-dominated regions. Effective polarization fusion should therefore treat $S_0$ and $DoLP$ as asymmetric yet complementary inputs rather than equally reliable modalities.

Against this background, recent multimodal image fusion research has progressively moved toward unified representation, stronger cross-modal interaction, generative modeling, and semantic guidance. EMMA introduces an equivariant self-supervised paradigm, while TC-MoA customizes shared representations for multiple fusion tasks through task-specific adapter routing~\cite{zhao2024emma,zhu2024tcmoa}. Task-aware Transformer interaction, high-order multimodal synergy, and discriminative evolutionary learning have also been explored to improve both fusion quality and downstream perception~\cite{chen2024hitfusion,zhou2025ship,liu2025dcevo}. Recent lightweight fusion models combine parallel Mamba--KAN modeling with heterogeneous multi-branch interaction, while adaptive multi-resolution enhancement and dynamic weighting improve scene-dependent information allocation~\cite{sun2025pmkfuse,wang2025multiresolution}. Diffusion-based frameworks such as FusionINV and LFDT-Fusion formulate multimodal integration through generative priors or latent diffusion Transformers~\cite{liang2025fusioninv,yang2025lfdt}. MaeFuse and OmniFuse further exploit pretrained visual representations and language-driven semantic guidance~\cite{li2025maefuse,zhang2025omnifuse}. Recent studies also address illumination-constrained fusion and compound adverse-weather restoration~\cite{peng2025bicfusion,liu2026cawmmamba}.

Despite these advances, their guidance is primarily derived from semantic relevance, downstream task utility, degradation characteristics, or learned data distributions. Polarization fusion poses a different problem because the physical reliability of $DoLP$ varies across spatial locations. Existing polarization-specific designs commonly apply priors, saliency cues, or attention mechanisms at selected stages, without a unified criterion for determining where and how strongly complementary polarization information should be transferred. Consequently, aggressive transfer may amplify unstable responses or disturb the luminance structure of $S_0$, whereas conservative regulation may suppress useful material- and reflection-sensitive details. Moreover, complex interaction or generative architectures introduce additional computational cost, while existing public polarization-fusion datasets remain limited in scale and scene diversity, as summarized in Table~\ref{table1}. These limitations motivate a lightweight framework that consistently regulates $DoLP$ transfer according to spatial confidence.

To address these issues, we propose LG-PF, a lightweight confidence-guided framework for polarization image fusion. LG-PF treats $S_0$ as the photometric and structural anchor and selectively transfers complementary $DoLP$ information through a residual process. The Polarization Confidence Prior (PCP) estimates the spatial reliability of polarization responses, the Mask-guided Multi-scale Fusion module (MMF) regulates polarization transfer across multiple feature scales, and the Lightweight Context-aware Bounded Correction Head (LCH) stabilizes local photometric and structural transitions. The same confidence prior is further incorporated into the optimization objectives to preserve reliable polarization details while suppressing unsupported responses.

The main contributions are summarized as follows:
\begin{enumerate}
\itemsep=0pt

\item We formulate polarization image fusion as a confidence-guided residual transfer problem, explicitly accounting for the spatially varying reliability of $DoLP$ information.

\item We develop a lightweight framework composed of PCP, MMF, and LCH for spatial confidence estimation, controlled multi-scale polarization transfer, and bounded local correction. The resulting model requires only $0.2936$ M parameters.

\item We construct MSP, a polarization fusion dataset containing 1000 pixel-aligned $S_0$--$DoLP$ image pairs from 17 indoor and outdoor scene categories. Extensive comparisons, cross-dataset evaluations, ablation studies, and efficiency analyses demonstrate the effectiveness, transferability, and computational efficiency of LG-PF.

\end{enumerate}

The remainder of this paper is organized as follows. Section~\ref{sec:related_work} reviews related studies. Section~\ref{sec:method} presents LG-PF. Section~\ref{sec:experiments} reports the experimental results, and Section~\ref{sec:conclusion} concludes the paper.

\section{Related Work}
\label{sec:related_work}

\subsection{Polarization Representation}
\label{sec:polarization_representation}

Polarization imaging records light-state information that is not available from intensity measurements alone. For the linear division-of-focal-plane (DoFP) imaging system considered in this work, four polarization observations, $I_{0^{\circ}}$, $I_{45^{\circ}}$, $I_{90^{\circ}}$, and $I_{135^{\circ}}$, are acquired, from which the Stokes parameters are computed~\cite{yang2024polarimetricreview,collett1984stokes}:
\begin{equation}
\begin{array}{c}
\displaystyle
S_0 = I_{0^{\circ}} + I_{90^{\circ}}
\\[3pt]
\displaystyle
S_1 = I_{0^{\circ}} - I_{90^{\circ}}
\\[3pt]
\displaystyle
S_2 = I_{45^{\circ}} - I_{135^{\circ}}
\end{array}
\label{eq:linear_stokes}
\end{equation}

The degree of linear polarization is calculated as
\begin{equation}
DoLP =
\frac{\sqrt{S_1^2+S_2^2}}{S_0+\epsilon},
\label{eq:dolp_definition}
\end{equation}
where $\epsilon$ is a small constant introduced to avoid numerical instability. Since $DoLP$ is normalized by $S_0$, its response may become less reliable in low-intensity, weakly polarized, or noise-dominated regions. Accordingly, $S_0$ and $DoLP$ should be regarded as asymmetric yet complementary modalities: $S_0$ provides a relatively stable photometric and structural reference, whereas $DoLP$ contributes material- and reflection-sensitive cues whose reliability varies spatially.

\subsection{Polarization Image Fusion Methods}
\label{sec:polarization_fusion_methods}

Based on this asymmetric yet complementary representation, polarization image fusion was initially formulated through multi-scale decomposition and rule-based coefficient selection. Oriented Laplacian pyramids, for example, integrate directional polarization information across multiple spatial scales~\cite{yue2014orientedlaplacian}. Although such methods offer clear physical interpretations, their performance is closely tied to the selected decomposition, activity measures, and reconstruction rules. Learning-based approaches subsequently shifted the focus toward adaptive representation and information weighting. Representative developments include self-learned $S_0$--$DoLP$ fusion~\cite{zhang2021selflearned}, quality- and attention-guided integration~\cite{duan2023dualweighted}, salient-information-guided fusion in PAPIF~\cite{xu2022papif}, and noise-aware filtering with a salient polarization prior~\cite{li2023noiseaware}. These methods established the importance of selectively emphasizing informative polarization responses rather than uniformly combining the two modalities.

Subsequent studies strengthened non-local interaction and modality-specific representation using deeper architectures. TIPFNet employs Transformer-based non-local modeling~\cite{li2022tipfnet}, whereas DT-F Transformer introduces dual-transpose interaction for cross-modal feature exchange~\cite{liu2024dtf}. PIPFNet incorporates polarization priors into feature extraction and reconstruction~\cite{li2024pipfnet}, and GAND adopts adversarial learning to enhance weak polarization targets~\cite{zhou2024gand}. CPIFuse further considers the physical discrepancy between intensity and polarization observations through lightweight color and texture reconstruction~\cite{luo2025cpifuse}. These approaches improve global dependency modeling and reconstruction flexibility, while embedding polarization regulation within different network components.

More recent work has pursued increasingly explicit polarization-aware modeling. Hierarchical progressive fusion uses significant polarization information to guide stage-wise integration~\cite{gong2025hierarchical}, while DCCFNet strengthens modality interaction through dual channel-cross fusion~\cite{liu2025dccfnet}. Beyond image-level reconstruction, FuseISP integrates intensity and spectral-polarization information through hierarchical multimodal fusion for transparent-object perception~\cite{fan2025fuseisp}. Polarization-forming-based fusion suppresses background interference before reconstruction~\cite{duan2025polarizationforming}, and IGFD-Net introduces illumination-guided frequency decoupling to balance polarization details and photometric fidelity~\cite{gong2026igfdnet}. PPA-Diff further extends polarization fusion toward physics-aware full-Stokes diffusion modeling~\cite{hu2026ppadiff}. Overall, polarization fusion has progressed from hand-crafted decomposition to global interaction and physics-aware representation. Nevertheless, most existing approaches embed polarization guidance within dedicated modules or processing stages, while a shared reliability estimate that consistently governs multi-scale transfer and subsequent correction remains comparatively underexplored.
\section{Proposed Method}
\label{sec:method}

\begin{figure*}
    \includegraphics[width=1\linewidth]{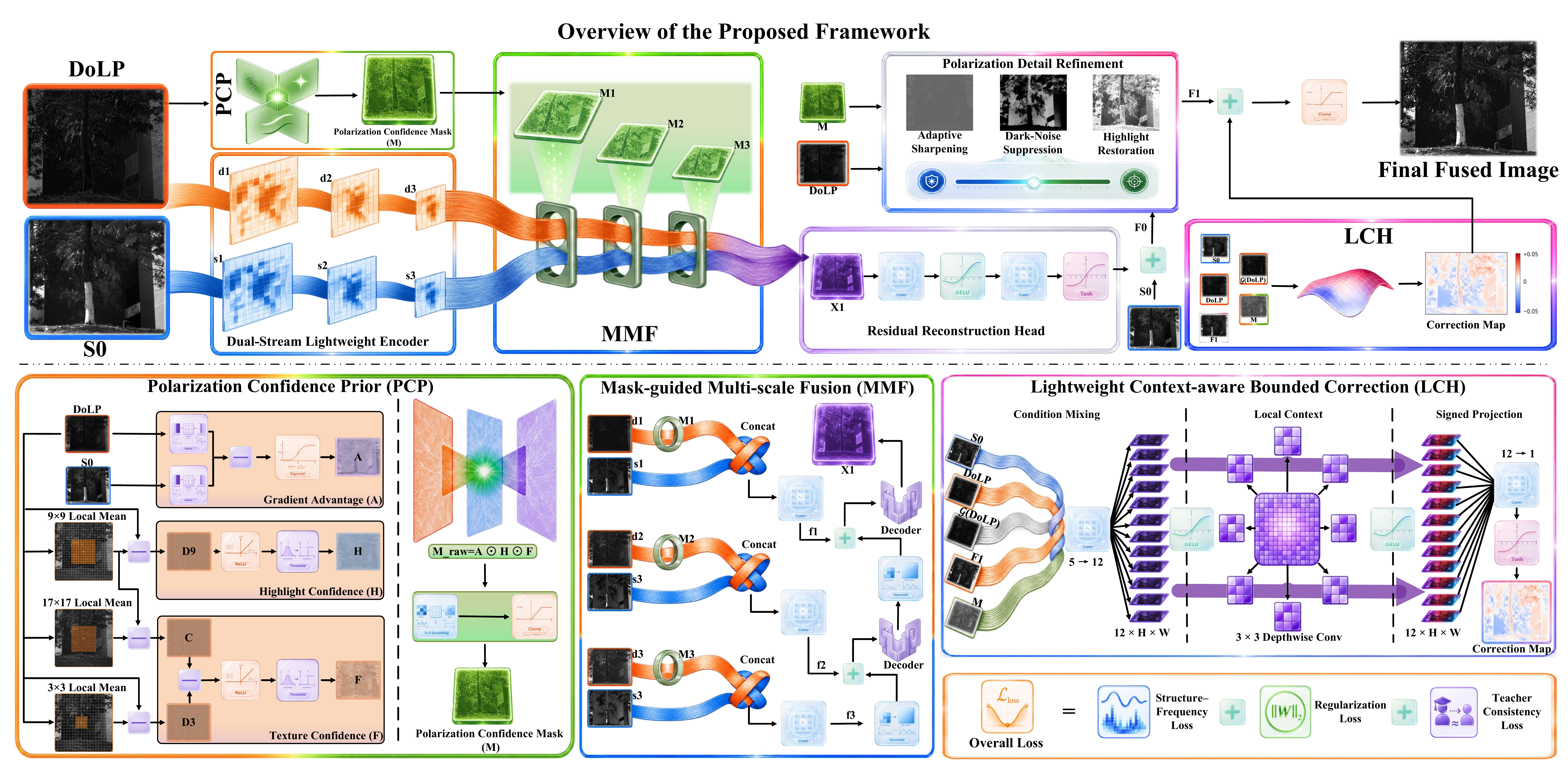}
    \vspace{-1em}
    \caption{Overall architecture of the proposed LG-PF.} 
    \vspace{-1.5em}
    \label{fig1}
\end{figure*}

\subsection{Overall Framework}

As illustrated in Fig.~\ref{fig1}, LG-PF instantiates the above reliability-guided formulation as a lightweight selective residual transfer framework. Given  $S_0$ and  $DoLP$, LG-PF treats $S_0$ as the photometric and structural anchor, while selectively transferring complementary polarization increments from $DoLP$.

The framework consists of three main components. The PCP estimates a polarization confidence mask $M$ from multi-scale local responses and the gradient advantage of $DoLP$ over $S_0$. The MMF resizes $M$ into scale-specific masks $\{M_i\}_{i=1}^{3}$ and uses them to regulate polarization feature transfer at three encoder stages. Finally, the LCH performs bounded local correction to local photometric and structural transitions in the fused result.

A lightweight dual-stream encoder with non-shared parameters extracts the intensity features $\{s_i\}_{i=1}^{3}$ and polarization features $\{d_i\}_{i=1}^{3}$. MMF produces the fused features $\{f_i\}_{i=1}^{3}$, which are progressively aggregated by the decoder to obtain the full-resolution representation $X_1$. A residual reconstruction head predicts a polarization increment relative to $S_0$, yielding the preliminary result $F_0$. Guided by $DoLP$ and $M$, the polarization detail refinement stage further restores reliable polarization-sensitive details to produce $F_1$. LCH then performs bounded local correction using $S_0$, $\mathcal{G}(DoLP)$, $DoLP$, $F_1$, and $M$, where $\mathcal{G}(\cdot)$ denotes the gradient-magnitude operator. The overall mapping is formulated as

\begin{equation}
\begin{array}{c}
\displaystyle
M=\mathcal{P}(S_0,DoLP)
\\[3pt]
\displaystyle
f_i=\mathcal{M}_i
\left(
\mathcal{E}_{S}^{i}(S_0),
\mathcal{E}_{D}^{i}(DoLP),
M_i
\right),
\quad i\in\{1,2,3\}
\\[3pt]
\displaystyle
X_1=
\mathcal{D}_{\mathrm{dec}}
\left(
f_1,f_2,f_3
\right)
\\[3pt]
\displaystyle
F_0=
\Pi_{[0,1]}
\left[
S_0+\lambda_r\mathcal{R}(X_1)
\right]
\\[3pt]
\displaystyle
F_1=
\mathcal{D}_{\mathrm{ref}}
\left(
F_0,DoLP,M
\right)
\\[3pt]
\displaystyle
I_{\mathrm{fuse}}=
\Pi_{[0,1]}
\left[
F_1+\lambda_c
\mathcal{C}
\left(
S_0,\mathcal{G}(DoLP),DoLP,F_1,M
\right)
\right]
\end{array}
\label{eq:overall_lgpf}
\end{equation}
Here, $\mathcal{P}$ denotes PCP, and $\mathcal{M}_i$ denotes the confidence-guided fusion operation at the $i$-th scale of MMF. $\mathcal{E}_{S}^{i}$ and $\mathcal{E}_{D}^{i}$ are the corresponding $S_0$ and $DoLP$ encoders, while $M_i$ is obtained by resizing $M$ to the spatial resolution of the $i$-th-scale features. $\mathcal{D}_{\mathrm{dec}}$, $\mathcal{R}$, and $\mathcal{D}_{\mathrm{ref}}$ represent the lightweight decoder, residual reconstruction head, and polarization detail refinement stage within MMF, respectively, whereas $\mathcal{C}$ denotes LCH. The coefficients $\lambda_r$ and $\lambda_c$ control the residual reconstruction and local correction magnitudes, respectively. $I_{\mathrm{fuse}}$ is the final fused image, and $\Pi_{[0,1]}(\cdot)$ projects its input onto the valid intensity range.
\subsection{PCP}
To realize the mapping $M=\mathcal{P}(S_0,DoLP)$ in Eq.~\eqref{eq:overall_lgpf}, PCP constructs a spatial confidence mask by jointly evaluating local response prominence, fine-scale texture evidence, and the cross-modal gradient advantage of $DoLP$ over $S_0$. Let $\mathcal{A}_k(\cdot)$ denote a $k\times k$ local-mean operator. Specifically, $D_9$ and $D_3$ are obtained by subtracting the $9\times9$ and $3\times3$ local-mean responses from $DoLP$, respectively, while $C$ is the difference between its $9\times9$ and $17\times17$ local-mean responses. Accordingly, $D_9$ characterizes locally prominent polarization responses, $D_3$ captures fine-scale variations, and $C$ represents coarse background fluctuations. Based on these responses, the highlight confidence $H$ and texture confidence $F$ are defined as
\begin{equation}
\begin{array}{c}
\displaystyle
H=
\sigma\!\left\{
\gamma_h
\left[
\operatorname{ReLU}(D_9)-\tau_h
\right]
\right\}
\\[3pt]
\displaystyle
F=
\sigma\!\left\{
\gamma_t
\left[
\operatorname{ReLU}
\left(
|D_3|-\alpha|C|
\right)
-\tau_t
\right]
\right\}
\end{array}
\label{eq:pcp_local_confidence}
\end{equation}
where $\sigma(\cdot)$ denotes the sigmoid function, $\operatorname{ReLU}(\cdot)$ retains positive responses, and $|\cdot|$ denotes the element-wise absolute value. $\gamma_h$ and $\gamma_t$ control the transition sharpness, $\tau_h$ and $\tau_t$ are the corresponding response thresholds, and $\alpha$ determines the suppression strength of coarse background variation. Thus, $H$ emphasizes locally prominent responses, whereas $F$ retains fine textures after suppressing large-scale fluctuations.

Local evidence from $DoLP$ alone cannot determine whether a response is complementary to the structural information in $S_0$. PCP therefore constructs a cross-modal gradient-advantage map:
\begin{equation}
A=
\beta_0+
\beta_1
\sigma
\left\{
\gamma_g
\left[
\mathcal{G}(DoLP)-\mathcal{G}(S_0)
\right]
\right\}
\label{eq:pcp_gradient_advantage}
\end{equation}
where $\beta_0$ sets the lower response bound, $\beta_1$ controls the adaptive range, and $\gamma_g$ determines the sensitivity to the cross-modal gradient difference. A larger value of $A$ indicates that the corresponding structure is more pronounced in $DoLP$ than in $S_0$ and is therefore assigned greater confidence during polarization transfer.

The three confidence cues are combined and locally smoothed to obtain the final polarization confidence mask:
\begin{equation}
\begin{array}{c}
\displaystyle
M_{\mathrm{raw}}=A\odot H\odot F
\\[3pt]
\displaystyle
M=
\Pi_{[0,1]}
\left[
\mathcal{A}_3
\left(
M_{\mathrm{raw}}
\right)
\right]
\end{array}
\label{eq:pcp_texture_mask}
\end{equation}
here, $\odot$ denotes element-wise multiplication. The multiplicative formulation assigns high confidence only to responses that are locally prominent, contain reliable fine-scale evidence, and provide a gradient advantage over $S_0$. Local-mean filtering improves spatial continuity and suppresses isolated responses. The resulting mask $M$ is resized to $\{M_i\}_{i=1}^{3}$ and passed to MMF for multi-scale polarization feature modulation.

\subsection{MMF}

To realize the multi-scale feature transfer in Eq.~\eqref{eq:overall_lgpf}, a non-shared dual-stream encoder extracts three-scale intensity features $\{s_i\}_{i=1}^{3}$ and polarization features $\{d_i\}_{i=1}^{3}$. Their channel dimensions are $c_i\in\{16,32,64\}$ for $i=1,2,3$, respectively. The confidence mask $M$ is resized to the spatial resolution of each feature scale, yielding the scale-specific masks $\{M_i\}_{i=1}^{3}$.

At each scale, MMF regulates the polarization feature using a soft confidence gate:
\begin{equation}
\widetilde{d}_i
=
\left[
\eta+(1-\eta)M_i
\right]\odot d_i,
\quad i\in\{1,2,3\}
\label{eq:mmf_gating}
\end{equation}
where $\widetilde{d}_i$ denotes the gated polarization feature and $\eta=0.35$ is the minimum retention coefficient. Since $M_i\in[0,1]$, the modulation weight lies in $[0.35,1]$, preserving high-confidence responses while attenuating, rather than completely discarding, uncertain polarization information.

The gated polarization feature is subsequently concatenated with the corresponding intensity feature and fused through a $1\times1$ convolution:
\begin{equation}
\begin{array}{c}
\displaystyle
z_i=
\operatorname{Cat}
\left(
s_i,\widetilde{d}_i
\right)
\\[3pt]
\displaystyle
f_i=
\operatorname{Conv}_{1\times1}^{2c_i\rightarrow c_i}
\left(
z_i
\right),
\quad i\in\{1,2,3\}
\end{array}
\label{eq:mmf_fusion}
\end{equation}
where $\operatorname{Cat}(\cdot)$ denotes channel-wise concatenation, and $z_i$ is the concatenated feature at the $i$-th scale. $\operatorname{Conv}_{1\times1}^{2c_i\rightarrow c_i}(\cdot)$ performs cross-modal channel mixing while reducing the channel dimension from $2c_i$ to $c_i$. The resulting $f_i$ corresponds to the output of $\mathcal{M}_i$ in Eq.~\eqref{eq:overall_lgpf}.

The fused features are then aggregated by a two-stage lightweight decoder. Starting from the deepest feature $f_3$, the decoder progressively upsamples and combines it with $f_2$ and $f_1$ to obtain the full-resolution representation $X_1$. The residual reconstruction head $\mathcal{R}(\cdot)$ maps $X_1$ to a single-channel polarization increment. With $\lambda_r=0.60$, this increment is added to the structural anchor $S_0$ to produce the preliminary fused result $F_0$. This residual formulation preserves the global appearance of $S_0$ while focusing the network on complementary polarization information.

Finally, the polarization detail refinement stage $\mathcal{D}_{\mathrm{ref}}$ uses $DoLP$ and $M$ to selectively restore reliable polarization-sensitive details that may be attenuated during hierarchical encoding and decoding. It produces the refined result $F_1$, which is subsequently passed to LCH for bounded local correction.
\subsection{LCH}
As the final stage in Eq.~\eqref{eq:overall_lgpf}, LCH performs lightweight bounded local correction on the refined result $F_1$, rather than reconstructing the fused image. It is conditioned on $S_0$, $\mathcal{G}(DoLP)$, $DoLP$, $F_1$, and $M$, which provide the structural reference, polarization-sensitive boundaries, original polarization responses, current fusion state, and spatial confidence guidance, respectively. These five single-channel maps are concatenated along the channel dimension to form the condition tensor $X_c$.

LCH first projects $X_c$ into a compact twelve-channel representation through a pointwise convolution, models local spatial context using a $3\times3$ depthwise convolution, and then predicts a signed single-channel correction:

\begin{equation}
\begin{array}{c}
\displaystyle
Z_c=
\delta
\left[
\operatorname{DWConv}_{3\times3}
\left(
\delta
\left[
\operatorname{Conv}_{1\times1}^{5\rightarrow12}
\left(
X_c
\right)
\right]
\right)
\right]
\\[3pt]
\displaystyle
\Delta_c=
\tanh
\left[
\operatorname{Conv}_{1\times1}^{12\rightarrow1}
\left(
Z_c
\right)
\right]
\end{array}
\label{eq:lch_correction}
\end{equation}
where $Z_c$ denotes the locally contextualized twelve-channel feature, $\delta(\cdot)$ represents the GELU activation, and $\operatorname{DWConv}_{3\times3}(\cdot)$ denotes the $3\times3$ depthwise convolution. The two pointwise projections perform cross-condition channel mixing and signed correction prediction, respectively. The hyperbolic tangent function constrains $\Delta_c$ to $[-1,1]$, enabling both positive and negative local adjustments while preventing unbounded modification.

The correction map $\Delta_c$ corresponds to the output of $\mathcal{C}(S_0,\mathcal{G}(DoLP),DoLP,F_1,M)$ in Eq.~\eqref{eq:overall_lgpf}. With $\lambda_c=0.05$, the maximum adjustment introduced by LCH is restricted to $\pm0.05$. This bounded correction stabilizes local photometric and structural transitions while preserving the structural and polarization information in $F_1$. Moreover, the pointwise projections and depthwise local modeling introduce only minor computational overhead.
\begin{table*}[t]
\centering
{\rmfamily
\caption{Comparison of representative open-source polarization fusion datasets. Boldface indicates the largest value in each count-based column, and N/R denotes information that was not reported.}
\label{table1}
\renewcommand{\arraystretch}{1.15}
\setlength{\tabcolsep}{6.5pt}
\begin{tabular}{ccccccc}
\hline
\multirow{2}{*}{Dataset}
& \multicolumn{6}{c}{Dataset details} \\
\cline{2-7}
& Acquisition pattern
& Total pairs
& Indoor pairs
& Outdoor pairs
& \shortstack{Scene\\categories}
& Image size \\
\hline

PIF~\cite{liu2024dtf}
& DoFP
& 74
& 5
& 69
& N/R
& $1024\times1224$ \\

GAND~\cite{zhou2024gand}
& DoFP
& 415
& \textbf{331}
& 84
& N/R
& $768\times576$ \\

\textbf{MSP (Proposed)}
& DoFP
& \textbf{1000}
& 128
& \textbf{872}
& \textbf{17}
& $1125\times938$ \\
\hline
\end{tabular}
}
\end{table*}

\subsection{Loss Functions}
\label{sec:loss_functions}

To optimize the above fusion pipeline without pixel-wise ground truth, LG-PF employs structure-frequency, regularization, and teacher consistency losses:
\begin{equation}
\mathcal{L}_{\mathrm{total}}
=
\lambda_{\mathrm{sf}}\mathcal{L}_{\mathrm{SF}}
+
\lambda_{\mathrm{reg}}\mathcal{L}_{\mathrm{REG}}
+
\lambda_{\mathrm{tc}}\mathcal{L}_{\mathrm{TC}}
\label{eq:total_loss}
\end{equation}
where $\lambda_{\mathrm{sf}}$, $\lambda_{\mathrm{reg}}$, and $\lambda_{\mathrm{tc}}$ balance the three loss groups. Their values are reported in the implementation details. All terms are applied to the final output $I_{\mathrm{fuse}}$.

Following PCP, $\mathcal{A}_k(\cdot)$ denotes $k\times k$ local-mean filtering, and
$\mathcal{H}_k(X)=X-\mathcal{A}_k(X)$ denotes the corresponding high-frequency component. Moreover, $\|\cdot\|_1$ is the $\ell_1$ norm, and $[x]_{+}=\max(x,0)$ is applied element-wise.

Using the confidence mask $M$, we construct the gradient and detail targets:
\begin{equation}
\begin{array}{c}
\displaystyle
G^{\star}
=
\max
\left[
\mathcal{G}(S_0),
M\odot\mathcal{G}(DoLP)
\right]
\\[4pt]
\displaystyle
D_k^{\star}
=
(1-M)\odot\mathcal{H}_k(S_0)
+
M\odot\mathcal{H}_k(DoLP),
\quad k\in\{3,9\}
\end{array}
\label{eq:loss_targets}
\end{equation}
where the maximum is applied element-wise. These targets preserve the structural information of $S_0$ while introducing polarization gradients and details according to the spatial confidence.

The structure-frequency loss is defined as
\begin{equation}
\mathcal{L}_{\mathrm{SF}}
=
\mathcal{L}_{\mathrm{grad}}
+
\mathcal{L}_{\mathrm{detail}}
+
\mathcal{L}_{\mathrm{pol}}
+
\mathcal{L}_{\mathrm{freq}}
\label{eq:structure_frequency_loss}
\end{equation}
with
\begin{equation}
\begin{array}{c}
\displaystyle
\mathcal{L}_{\mathrm{grad}}
=
\left\|
\left[
G^{\star}
-
\mathcal{G}(I_{\mathrm{fuse}})
\right]_{+}
\right\|_1
\\[5pt]
\displaystyle
\mathcal{L}_{\mathrm{detail}}
=
\left\|
\mathcal{H}_3(I_{\mathrm{fuse}})
-
D_3^{\star}
\right\|_1
\\[5pt]
\displaystyle
\mathcal{L}_{\mathrm{pol}}
=
\left\|
M\odot
\left[
\mathcal{H}_3(I_{\mathrm{fuse}})
-
\mathcal{H}_3(DoLP)
\right]
\right\|_1
\\[5pt]
\displaystyle
\mathcal{L}_{\mathrm{freq}}
=
\left\|
\mathcal{A}_9(I_{\mathrm{fuse}})
-
\mathcal{A}_9(S_0)
\right\|_1
+
\left\|
\mathcal{H}_9(I_{\mathrm{fuse}})
-
D_9^{\star}
\right\|_1
\end{array}
\label{eq:structure_frequency_components}
\end{equation}
here, $\mathcal{L}_{\mathrm{grad}}$ penalizes missing structural gradients, $\mathcal{L}_{\mathrm{detail}}$ constrains confidence-weighted fine details, and $\mathcal{L}_{\mathrm{pol}}$ further reinforces polarization consistency in high-confidence regions. $\mathcal{L}_{\mathrm{freq}}$ preserves the low-frequency appearance of $S_0$ and the confidence-guided intermediate-scale details.

To suppress excessive responses in dark regions and undesirable exposure variations, we define
\begin{equation}
\begin{array}{c}
\displaystyle
M_{\mathrm{dark}}
=
\sigma
\left[
\kappa_{\mathrm{d}}
\left(
\tau_{\mathrm{d}}
-
\mathcal{A}_{17}(S_0)
\right)
\right]
\\[5pt]
\displaystyle
E_{\mathrm{dark}}
=
\left|
\mathcal{H}_3(I_{\mathrm{fuse}})
\right|
-
\left|
\mathcal{H}_3(S_0)
\right|
-
M\odot
\left|
\mathcal{H}_3(DoLP)
\right|
\end{array}
\label{eq:dark_regularization_terms}
\end{equation}
where $\kappa_{\mathrm{d}}$ controls the transition sharpness and $\tau_{\mathrm{d}}$ is the dark-region threshold. $M_{\mathrm{dark}}$ emphasizes locally dark regions, while $E_{\mathrm{dark}}$ measures high-frequency responses exceeding the evidence provided by $S_0$ and confidence-weighted $DoLP$ details.

The regularization loss is
\begin{equation}
\mathcal{L}_{\mathrm{REG}}
=
\mathcal{L}_{\mathrm{dark}}
+
\mathcal{L}_{\mathrm{exp}}
\label{eq:regularization_loss}
\end{equation}
with
\begin{equation}
\begin{array}{c}
\displaystyle
\mathcal{L}_{\mathrm{dark}}
=
\left\|
M_{\mathrm{dark}}
\odot
\left[
E_{\mathrm{dark}}
\right]_{+}
\right\|_1
\\[5pt]
\displaystyle
\mathcal{L}_{\mathrm{exp}}
=
\left\|
\left[
\tau_{\mathrm{l}}
-
\mathcal{A}_{31}(I_{\mathrm{fuse}})
\right]_{+}
\right\|_1
+
\left\|
\left[
\mathcal{A}_{31}(I_{\mathrm{fuse}})
-
\tau_{\mathrm{u}}
\right]_{+}
\right\|_1
\end{array}
\label{eq:regularization_components}
\end{equation}
where $\tau_{\mathrm{l}}$ and $\tau_{\mathrm{u}}$ define the valid local-exposure range. $\mathcal{L}_{\mathrm{dark}}$ suppresses unsupported high-frequency responses in dark regions, whereas $\mathcal{L}_{\mathrm{exp}}$ penalizes local means outside the prescribed range.

Two frozen fusion teachers are further used to provide complementary supervision. Let $T_{\mathrm{P}}$ and $T_{\mathrm{L}}$ denote the outputs of PAPIF and LFDT, respectively. For $q\in\{\mathrm{P},\mathrm{L}\}$,
\begin{equation}
\begin{array}{c}
\displaystyle
\mathcal{L}_{q,\mathrm{low}}
=
\left\|
\mathcal{A}_{17}(I_{\mathrm{fuse}})
-
\mathcal{A}_{17}(T_q)
\right\|_1
\\[5pt]
\displaystyle
\mathcal{L}_{q,\mathrm{ssim}}
=
1-
\operatorname{SSIM}
\left(
\Psi_q(I_{\mathrm{fuse}}),
\Psi_q(T_q)
\right)
\\[5pt]
\displaystyle
\mathcal{L}_{\mathrm{TC}}
=
\sum_{q\in\{\mathrm{P},\mathrm{L}\}}
\left(
\mathcal{L}_{q,\mathrm{low}}
+
\mathcal{L}_{q,\mathrm{ssim}}
\right)
\end{array}
\label{eq:teacher_consistency_loss}
\end{equation}
where $\Psi_{\mathrm{P}}(X)=\mathcal{A}_{17}(X)$ and $\Psi_{\mathrm{L}}(X)=X$. PAPIF mainly provides low-frequency structural guidance, whereas LFDT additionally constrains the full-resolution appearance. Both teachers remain frozen during training and are removed during inference.
\section{Experiments}
\label{sec:experiments}

\subsection{MSP Dataset}
\label{sec:msp_dataset}

\begin{figure*}[t]
    \centering
    \includegraphics[width=\textwidth]{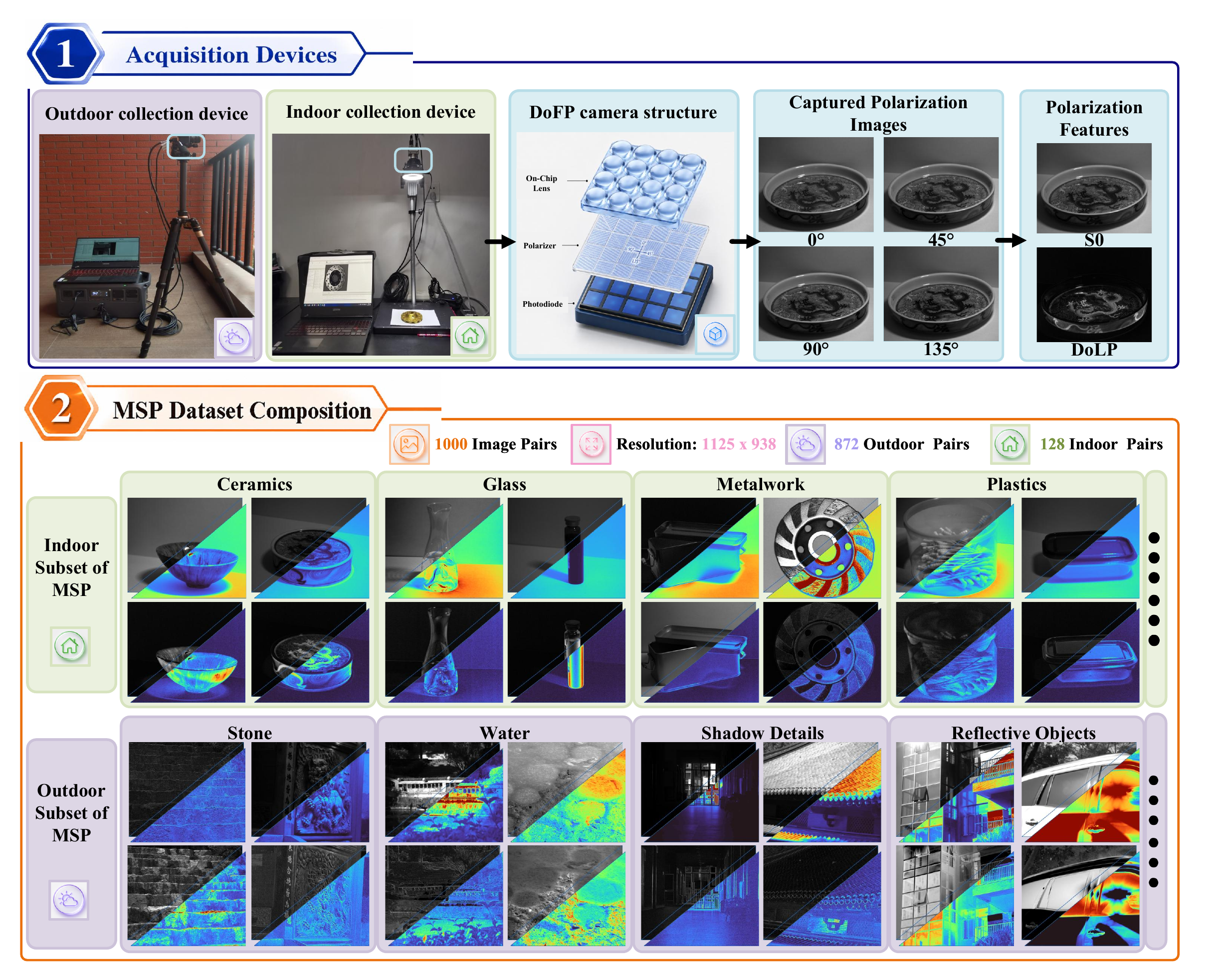}
    \caption{Acquisition systems and composition of the MSP dataset, including the DoFP imaging process, four-direction polarization observations, derived $S_0$ and $DoLP$ images, and representative indoor and outdoor scenes.}
    \label{fig:msp_dataset}
\end{figure*}

To support polarization image fusion in diverse real-world scenes, we construct a multi-scene polarization dataset termed MSP. As shown in Fig.~\ref{fig:msp_dataset}, MSP contains 1000 pixel-aligned $S_0$--$DoLP$ image pairs, including 872 outdoor pairs and 128 indoor pairs, acquired using division-of-focal-plane (DoFP) polarization imaging systems.

In a DoFP sensor, micro-polarizers at $0^{\circ}$, $45^{\circ}$, $90^{\circ}$, and $135^{\circ}$ are periodically integrated above the photodiode array. The four directional polarization observations are separated from each raw image and spatially interpolated, from which $S_0$ and $DoLP$ are derived according to the Stokes representation.

MSP covers diverse polarization-sensitive materials and imaging conditions. The indoor subset mainly includes ceramics, glass, metalwork, and plastics, whereas the outdoor subset contains stone surfaces, water regions, shadow details, and reflective objects. It further spans daytime and nighttime scenes, varying illumination and exposure levels, reflections, shadows, and a wide range of real-world environments.

Table~\ref{table1} compares MSP with the representative PIF~\cite{liu2024dtf} and GAND~\cite{zhou2024gand} datasets. Compared with these benchmarks, which contain fewer image pairs and are concentrated on a relatively limited set of acquisition scenes, MSP provides substantially broader coverage in terms of scene diversity, material categories, and illumination conditions. This diversity makes MSP a more comprehensive benchmark for training and evaluating polarization image fusion methods under real-world conditions.
\subsection{Experimental Settings}
\label{sec:experimental_settings}

The 1000 image pairs in MSP were randomly divided into 900 training pairs, 50 validation pairs, and 50 test pairs. The paired $S_0$ and $DoLP$ images were normalized to $[0,1]$ and resized to a common resolution of $1125\times938$ using bicubic interpolation. LG-PF was trained from scratch using AdamW with a batch size of 16, an initial learning rate of $5\times10^{-5}$, and a weight decay of $5\times10^{-5}$. A cosine annealing schedule reduced the learning rate to $2.5\times10^{-6}$. Training was conducted for 175 epochs, and the checkpoint obtained at epoch 130 was selected according to its validation performance. The maximum gradient norm was set to 0.5, the base channel width was 16, and four data-loading workers were used. All experiments were conducted on a single NVIDIA GeForce RTX 3090 GPU.

For cross-dataset evaluation, fixed subsets of 10 image pairs were randomly sampled from PIF~\cite{liu2024dtf} and GAND~\cite{zhou2024gand}, respectively. The same subsets were used for all compared methods, and LG-PF was directly evaluated without fine-tuning or domain adaptation.

LG-PF was compared with six representative methods: CPIFuse~\cite{luo2025cpifuse}, DT-F~\cite{liu2024dtf}, LFDT~\cite{yang2025lfdt}, PAPIF~\cite{xu2022papif}, PIPFNet~\cite{li2024pipfnet}, and TIPFNet~\cite{li2022tipfnet}. These methods cover lightweight CNN, attention-guided, Transformer-based, and polarization-prior-guided fusion paradigms. All fused results were converted to the same grayscale intensity range and evaluated using identical metric implementations.

Six commonly used fusion metrics were adopted: entropy (EN), spatial frequency (SF), standard deviation (SD) ~\cite{jagalingam2015review}, the sum of correlations of differences (SCD)~\cite{aslantas2015scd}, multi-scale structural similarity (MS-SSIM)~\cite{wang2003msssim}, and the Chen--Blum metric $Q_{CB}$~\cite{chen2009qcb}. EN, SF, and SD measure information content, spatial detail, and contrast, respectively. SCD evaluates complementary information transfer, whereas MS-SSIM and $Q_{CB}$ assess structural and perceptual quality. Higher values indicate better performance for all six metrics.
\subsection{Comparison Results and Analysis}

\label{sec:comparison_results}
\begin{figure*}
\includegraphics[width=1\linewidth]{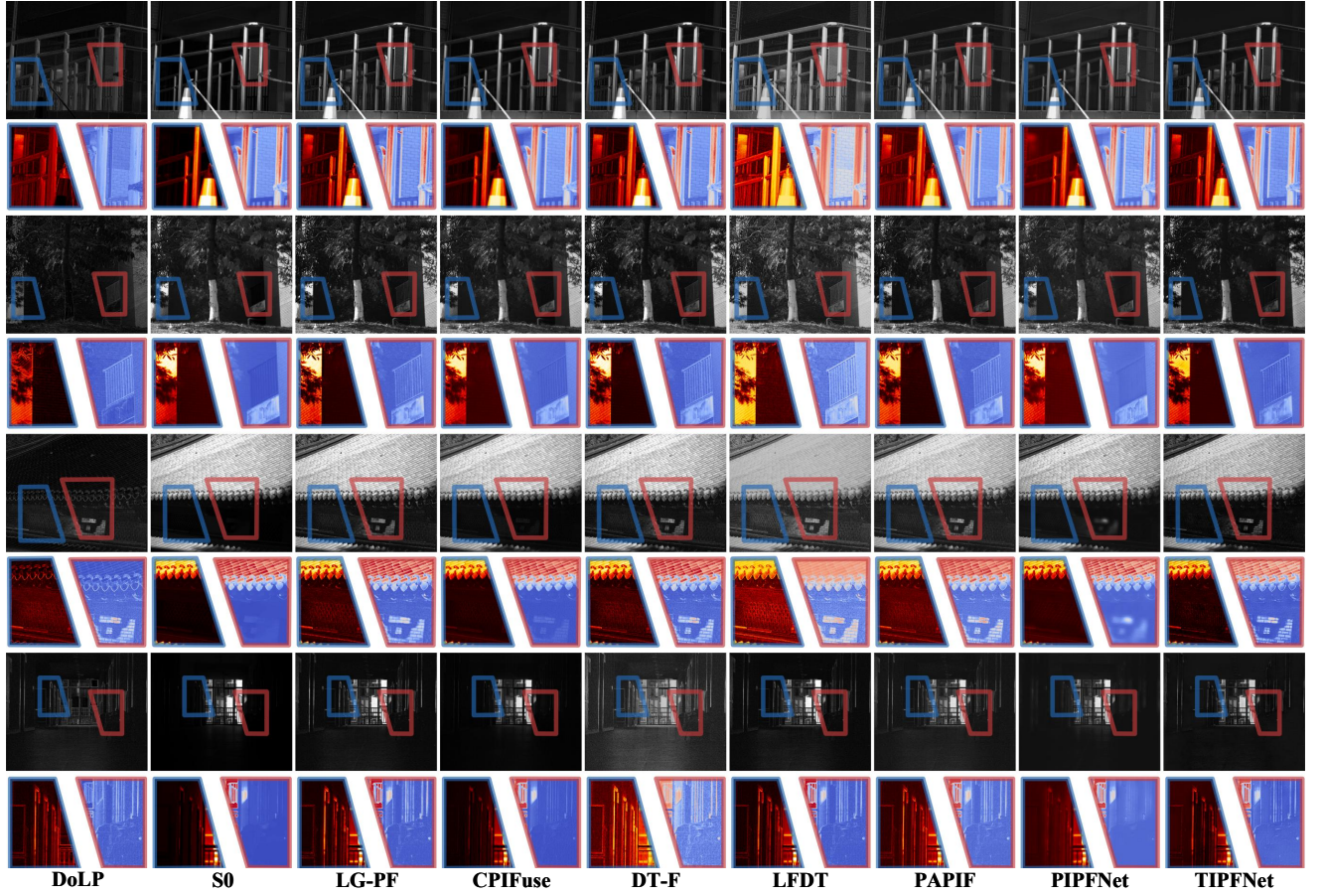}
\caption{Qualitative comparison on representative samples from the MSP test set.}
\label{fig:msp_visual}
\vspace{-1.5em}
\end{figure*}
\begin{table}[!t]
\centering
\small
\setlength{\tabcolsep}{3pt}
{\rmfamily
\caption{Quantitative comparison results on the MSP test set. Red and blue denote the best and second-best results, respectively.}
\label{tab:msp_results}
\resizebox{\columnwidth}{!}{
\begin{tabular}{ccccccc}
\toprule
\multirow{2}{*}{Methods} &
\multicolumn{6}{c}{Metrics} \\
\cmidrule(lr){2-7}
& EN$\uparrow$
& SF$\uparrow$
& SD$\uparrow$
& SCD$\uparrow$
& MS-SSIM$\uparrow$
& $Q_{CB}\uparrow$ \\
\midrule
\textbf{LG-PF}
& {\color[HTML]{FE0000}6.5485}
& {\color[HTML]{FE0000}15.6979}
& {\color[HTML]{FE0000}46.0210}
& {\color[HTML]{FE0000}1.6632}
& {\color[HTML]{FE0000}0.9439}
& {\color[HTML]{FE0000}0.3099} \\

CPIFuse
& 6.3884
& 8.9657
& 40.8085
& 1.4867
& 0.9296
& 0.3020 \\

DT-F
& {\color[HTML]{34CDF9}6.5269}
& {\color[HTML]{34CDF9}15.5525}
& 40.4660
& {\color[HTML]{34CDF9}1.6563}
& 0.7926
& 0.2621 \\

LFDT
& 6.5263
& 12.1124
& {\color[HTML]{34CDF9}45.5976}
& 1.6251
& {\color[HTML]{34CDF9}0.9387}
& {\color[HTML]{34CDF9}0.3074} \\

PAPIF
& 6.4486
& 10.8050
& 42.7253
& 1.5935
& 0.9057
& 0.2892 \\

PIPFNet
& 6.3281
& 8.0722
& 39.0078
& 1.1774
& 0.7650
& 0.2727 \\

TIPFNet
& 6.3078
& 10.8383
& 38.0155
& 1.5136
& 0.8252
& 0.2996 \\
\bottomrule
\end{tabular}
}
}
\end{table}

Tables~\ref{tab:msp_results}--\ref{tab:gand_results} summarize the quantitative comparisons on the full MSP test set and the fixed PIF and GAND subsets. The best and second-best values are highlighted in red and blue, respectively. As shown in Table~\ref{tab:msp_results}, LG-PF achieves the best performance across all six metrics on MSP. Its leading EN, SF, and SD values indicate strong information preservation, spatial detail, and contrast, while the superior SCD, MS-SSIM, and $Q_{CB}$ scores demonstrate effective complementary information transfer, structural fidelity, and perceptual quality. These results show that LG-PF improves polarization-sensitive details without sacrificing the photometric and structural information anchored by $S_0$.

\begin{figure*}
\includegraphics[width=1\linewidth]{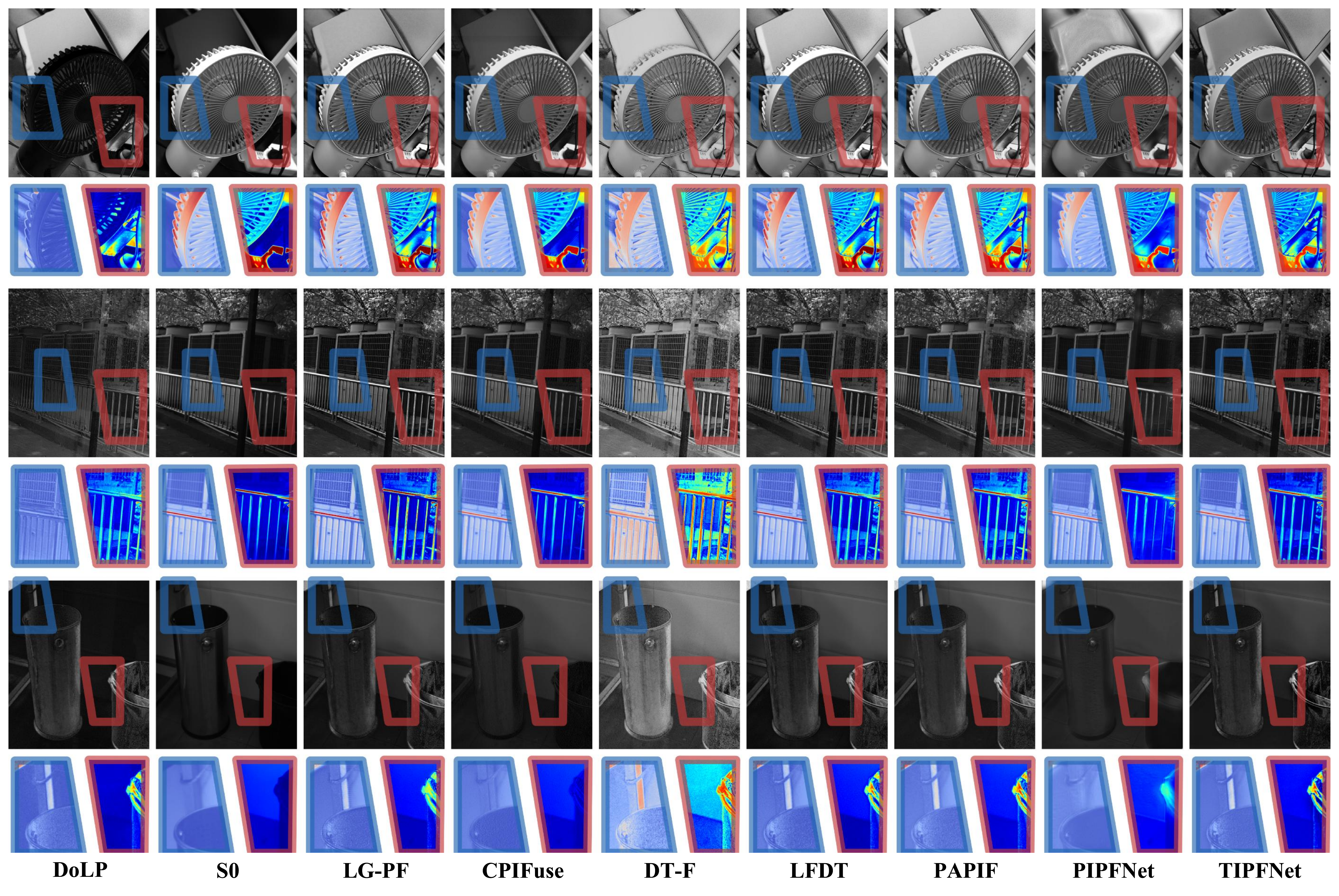}
\caption{Qualitative comparison on randomly selected PIF samples without fine-tuning. }
\label{fig:pif_visual}
\vspace{-1.5em}
\label{fig5}
\end{figure*}

\begin{table}[!t]
\centering
\small
\setlength{\tabcolsep}{3pt}
{\rmfamily
\caption{Quantitative comparison results on 10 randomly selected image pairs from the PIF dataset. }
\label{tab:pif_results}
\resizebox{\columnwidth}{!}{
\begin{tabular}{ccccccc}
\toprule
\multirow{2}{*}{Methods} &
\multicolumn{6}{c}{Metrics} \\
\cmidrule(lr){2-7}
& EN$\uparrow$
& SF$\uparrow$
& SD$\uparrow$
& SCD$\uparrow$
& MS-SSIM$\uparrow$
& $Q_{CB}\uparrow$ \\
\midrule
\textbf{LG-PF}
& {\color[HTML]{FE0000}7.4989}
& {\color[HTML]{FE0000}23.2688}
& {\color[HTML]{FE0000}63.8530}
& 1.5917
& {\color[HTML]{FE0000}0.9214}
& {\color[HTML]{34CDF9}0.4562} \\

CPIFuse
& 7.3479
& 13.8353
& 55.0505
& 1.5699
& 0.9155
& 0.4165 \\

DT-F
& 7.3064
& {\color[HTML]{34CDF9}21.7685}
& 52.4917
& {\color[HTML]{FE0000}1.7609}
& 0.8166
& 0.3603 \\

LFDT
& {\color[HTML]{34CDF9}7.4509}
& 17.8186
& {\color[HTML]{34CDF9}62.7805}
& 1.5718
& {\color[HTML]{34CDF9}0.9175}
& {\color[HTML]{FE0000}0.4728} \\

PAPIF
& 7.4261
& 17.0269
& 60.0223
& {\color[HTML]{34CDF9}1.6770}
& 0.9140
& 0.4455 \\

PIPFNet
& 7.2163
& 13.1146
& 55.7901
& 1.4345
& 0.8443
& 0.3887 \\

TIPFNet
& 7.2585
& 15.7598
& 52.8156
& 1.5841
& 0.8583
& 0.4494 \\
\bottomrule
\end{tabular}
}
}
\end{table}

The visual comparisons in Fig.~\ref{fig:msp_visual} are consistent with the quantitative results. In the first and fourth examples, the red-box regions show that LG-PF preserves reliable polarization-sensitive details in dark areas, whereas the blue-box regions demonstrate that the luminance and structural content of $S_0$ remains well maintained. The second and third examples contain challenging bright--dark transitions. LG-PF retains fine wall-surface textures while recovering polarization details from dark iron railings and tiled surfaces. In contrast, several competing methods either attenuate informative local details, introduce excessive polarization enhancement, or disturb the original luminance distribution. LG-PF therefore achieves a favorable balance between selective $DoLP$ detail transfer and preservation of the photometric and structural appearance of $S_0$.

On the fixed PIF subset, Table~\ref{tab:pif_results} shows that LG-PF ranks first in EN, SF, SD, and MS-SSIM, and second in $Q_{CB}$. As illustrated in Fig.~\ref{fig:pif_visual}, LG-PF more clearly preserves radial fan textures, object boundaries, and material-sensitive details, without introducing obvious luminance deviation or local over-enhancement.

\begin{figure*}
\includegraphics[width=1\linewidth]{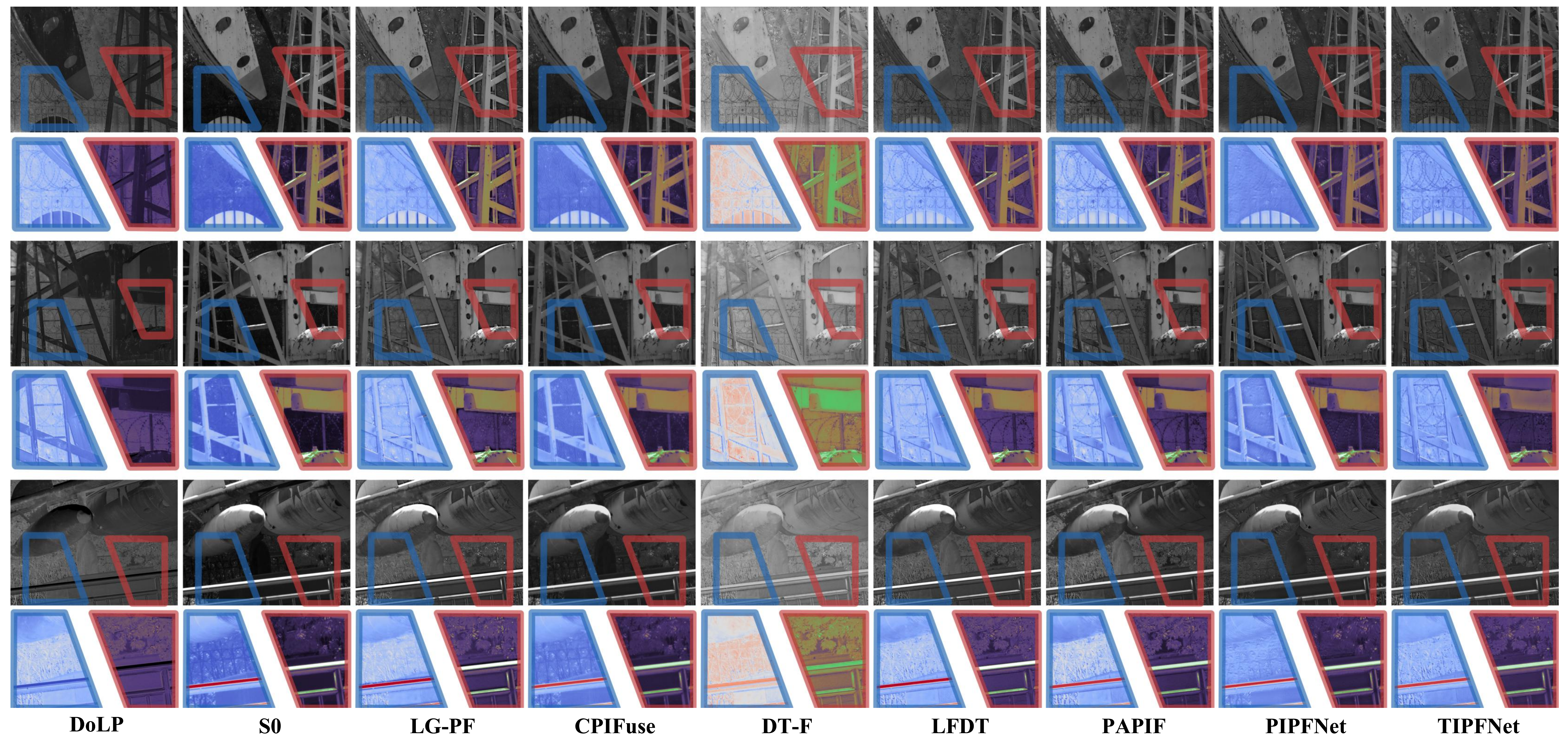}
\caption{Qualitative comparison on a randomly selected GAND sample without fine-tuning. }
\label{fig:gand_visual}
\vspace{-1.5em}
\end{figure*}

On the fixed GAND subset, LG-PF obtains the best EN, SF, and SD values and the second-best $Q_{CB}$ value, as reported in Table~\ref{tab:gand_results}. Although DT-F achieves a higher SCD score and CPIFuse performs better in MS-SSIM and $Q_{CB}$, Fig.~\ref{fig:gand_visual} shows that LG-PF retains clear scene structures, dark-region details, and polarization-sensitive boundaries while maintaining a natural intensity distribution. The evaluations on PIF and GAND therefore indicate promising cross-dataset transferability without fine-tuning or domain adaptation.

\begin{table}[!t]
\centering
\small
\setlength{\tabcolsep}{3pt}
{\rmfamily
\caption{Quantitative comparison results on 10 randomly selected image pairs from the GAND dataset. }
\label{tab:gand_results}
\resizebox{\columnwidth}{!}{
\begin{tabular}{ccccccc}
\toprule
\multirow{2}{*}{Methods} &
\multicolumn{6}{c}{Metrics} \\
\cmidrule(lr){2-7}
& EN$\uparrow$
& SF$\uparrow$
& SD$\uparrow$
& SCD$\uparrow$
& MS-SSIM$\uparrow$
& $Q_{CB}\uparrow$ \\
\midrule
\textbf{LG-PF}
& {\color[HTML]{FE0000}6.9700}
& {\color[HTML]{FE0000}17.6705}
& {\color[HTML]{FE0000}39.1293}
& 1.7269
& 0.7335
& {\color[HTML]{34CDF9}0.5290} \\

CPIFuse
& 6.8694
& 11.3982
& 36.9481
& 1.7230
& {\color[HTML]{FE0000}0.9430}
& {\color[HTML]{FE0000}0.5883} \\

DT-F
& 6.8655
& {\color[HTML]{34CDF9}16.6782}
& 29.9436
& {\color[HTML]{FE0000}1.8802}
& 0.6180
& 0.3696 \\

LFDT
& 6.8695
& 14.7905
& 38.2868
& 1.7196
& 0.5697
& 0.4923 \\

PAPIF
& {\color[HTML]{34CDF9}6.9613}
& 14.5522
& {\color[HTML]{34CDF9}38.8881}
& 1.7233
& 0.6147
& 0.4716 \\

PIPFNet
& 6.8056
& 13.4443
& 36.9769
& 1.6107
& {\color[HTML]{34CDF9}0.8007}
& 0.5078 \\

TIPFNet
& 6.6308
& 12.8721
& 31.6894
& {\color[HTML]{34CDF9}1.8365}
& 0.7013
& 0.5261 \\
\bottomrule
\end{tabular}
}
}
\end{table}

Overall, LG-PF achieves the strongest quantitative performance on MSP and remains competitive on the two external datasets. The quantitative and qualitative results consistently support its ability to selectively transfer reliable polarization information while preserving structural fidelity and luminance consistency across different imaging conditions.
\subsection{Ablation Studies}
\label{sec:ablation}

All ablation variants were independently trained from scratch on the same MSP training split under identical settings, with only the investigated component or loss configuration changed. For ablation evaluation, fixed subsets were randomly sampled once from MSP, PIF, and GAND and consistently used for the full model and all ablation variants. The PIF and GAND subsets were evaluated without fine-tuning or domain adaptation.
\begin{table}[!t]
\centering
\small
\setlength{\tabcolsep}{3pt}
{\rmfamily
\caption{Ablation study of the key modules.  }
\label{table:module_ablation}
\resizebox{\columnwidth}{!}{
\begin{tabular}{lcccccc}
\toprule
\multirow{2}{*}{Methods} &
\multicolumn{6}{c}{Metrics} \\
\cmidrule(lr){2-7}
& EN$\uparrow$
& SF$\uparrow$
& SD$\uparrow$
& SCD$\uparrow$
& MS-SSIM$\uparrow$
& $Q_{CB}\uparrow$ \\
\midrule

Full model
& {\color[HTML]{FE0000}6.9094}
& 15.5907
& {\color[HTML]{34CDF9}56.3164}
& {\color[HTML]{FE0000}1.7211}
& {\color[HTML]{34CDF9}0.9466}
& 0.4165 \\

w/o PCP
& 6.8752
& {\color[HTML]{34CDF9}16.5595}
& {\color[HTML]{FE0000}56.3806}
& 1.7019
& 0.9251
& 0.4119 \\

w/o MMF
& 6.8931
& 15.3270
& 55.8327
& {\color[HTML]{34CDF9}1.7063}
& {\color[HTML]{FE0000}0.9475}
& {\color[HTML]{FE0000}0.4224} \\

w/o LCH
& {\color[HTML]{34CDF9}6.8962}
& 15.1074
& 56.2214
& 1.7054
& 0.9403
& {\color[HTML]{34CDF9}0.4190} \\

Baseline
& 6.8653
& {\color[HTML]{FE0000}16.8300}
& 56.1553
& 1.7012
& 0.9233
& 0.4129 \\

\bottomrule
\end{tabular}
}
}
\end{table}
Table~\ref{table:module_ablation} evaluates the contributions of PCP, MMF, and LCH. Removing PCP increases SF and SD but degrades EN, SCD, MS-SSIM, and $Q_{CB}$, indicating that stronger local activity may also introduce unreliable polarization responses. Removing MMF reduces information content and source complementarity, while removing LCH weakens photometric and structural consistency. Although the baseline achieves the highest SF, its lower structural and perceptual scores show that spatial activity alone does not guarantee balanced fusion quality. Overall, the complete model provides the most consistent performance across the six metrics.

\begin{table}[!t]
\centering
\small
\setlength{\tabcolsep}{3pt}
{\rmfamily
\caption{Ablation study of different loss configurations. }
\label{table:loss_ablation}
\resizebox{\columnwidth}{!}{
\begin{tabular}{ccccccc}
\toprule
\multirow{2}{*}{Settings} &
\multicolumn{6}{c}{Metrics} \\
\cmidrule(lr){2-7}
& EN$\uparrow$
& SF$\uparrow$
& SD$\uparrow$
& SCD$\uparrow$
& MS-SSIM$\uparrow$
& $Q_{CB}\uparrow$ \\
\midrule

Full loss
& {\color[HTML]{FE0000}6.9094}
& 15.5907
& 56.3164
& {\color[HTML]{FE0000}1.7211}
& {\color[HTML]{34CDF9}0.9466}
& {\color[HTML]{FE0000}0.4165} \\

w/o $\mathcal{L}_{\mathrm{REG}}$
& 6.8767
& {\color[HTML]{34CDF9}15.6931}
& 56.1723
& 1.6984
& {\color[HTML]{FE0000}0.9535}
& 0.4119 \\

w/o $\mathcal{L}_{\mathrm{SF}}$
& {\color[HTML]{34CDF9}6.8942}
& {\color[HTML]{FE0000}15.7950}
& {\color[HTML]{34CDF9}56.7184}
& {\color[HTML]{34CDF9}1.7093}
& 0.9230
& 0.4147 \\

w/o $\mathcal{L}_{\mathrm{TC}}$
& 6.7902
& 15.3528
& {\color[HTML]{FE0000}56.7781}
& 1.3409
& 0.8917
& 0.4107 \\

$\mathcal{L}_{\mathrm{TC}}$ with PAPIF only
& 6.8869
& 15.5807
& 55.4955
& 1.6954
& 0.9316
& 0.4005 \\

$\mathcal{L}_{\mathrm{TC}}$ with LFDT only
& 6.7906
& 14.9664
& 55.9281
& 1.6028
& 0.9415
& {\color[HTML]{34CDF9}0.4160} \\

\bottomrule
\end{tabular}
}
}
\end{table}

\begin{figure}
    \centering
    \includegraphics[width=\columnwidth]{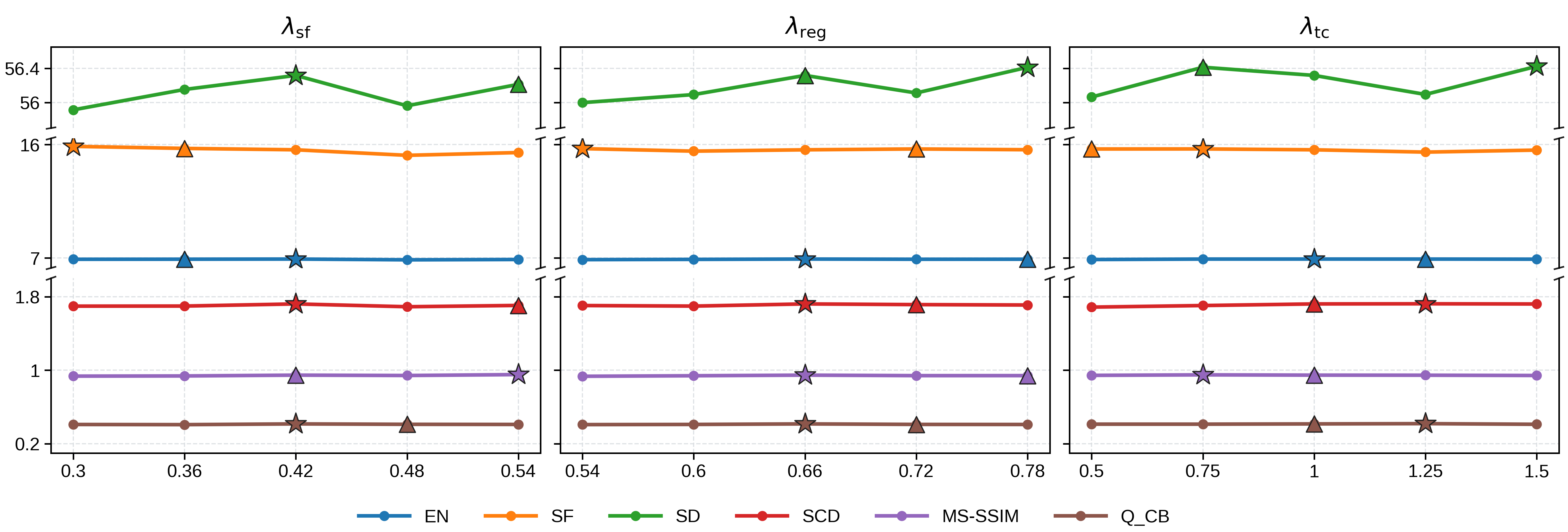}
    \caption{Sensitivity analysis of the three group-level loss
    weights. Stars and triangles denote the best and second-best
    results for each metric, respectively.}
    \label{fig:loss_sensitivity}
\end{figure}

Table~\ref{table:loss_ablation} evaluates the three loss groups and the teacher composition of $\mathcal{L}_{\mathrm{TC}}$. Removing $\mathcal{L}_{\mathrm{SF}}$ or $\mathcal{L}_{\mathrm{REG}}$ improves several individual metrics but weakens the overall balance, whereas removing $\mathcal{L}_{\mathrm{TC}}$ causes substantial degradation in EN, SCD, and MS-SSIM. Using PAPIF or LFDT alone also underperforms the complete configuration, confirming the complementary roles of structural and full-resolution appearance guidance.

Figure~\ref{fig:loss_sensitivity} further evaluates the sensitivity of the three group-level loss weights. The adopted setting, $\lambda_{\mathrm{sf}}=0.42$, $\lambda_{\mathrm{reg}}=0.66$, and $\lambda_{\mathrm{tc}}=1.00$, achieves the largest number of best-performing results and provides a favorable balance across the six metrics. The metric curves also remain relatively stable within the evaluated ranges, indicating that LG-PF is not highly sensitive to moderate variations in these weights.
\subsection{Computational Efficiency}
\label{sec:efficiency}

Table~\ref{table:efficiency} compares the computational costs of all methods under the same input resolution and hardware platform. LG-PF achieves the lowest computational complexity of 51.9883 G FLOPs and the fastest inference time of 21.7120 ms, while requiring only 0.2936 M parameters. Compared with the next-lowest computational cost achieved by PIPFNet, LG-PF reduces FLOPs by approximately 31.9\%. It also reduces inference time by approximately 80.4\% relative to the next-fastest method, CPIFuse.

Although PIPFNet and PAPIF contain fewer parameters, both require substantially more computation and longer inference time. This result indicates that parameter count alone does not fully reflect practical efficiency. The efficiency of LG-PF benefits from its compact multi-scale representation, channel-compressed fusion operations, residual reconstruction strategy, and depthwise spatial modeling in LCH. Combined with the quantitative results in Table~\ref{tab:msp_results}, these findings demonstrate that LG-PF achieves high-quality polarization fusion without relying on a parameter-heavy or computationally expensive architecture.

\begin{table}[!t]
\centering
\footnotesize
\setlength{\tabcolsep}{3pt}
\renewcommand{\arraystretch}{0.95}
{\rmfamily
\caption{Comparison of computational efficiency. }
\label{table:efficiency}
\resizebox{0.82\columnwidth}{!}{
\begin{tabular}{cccc}
\toprule
\multirow{2}{*}{Methods}
& \multicolumn{3}{c}{Computational efficiency} \\
\cmidrule(lr){2-4}
& FLOPs (G)$\downarrow$
& Parameters (M)$\downarrow$
& Time (ms)$\downarrow$ \\
\midrule

\textbf{LG-PF}
& {\color[HTML]{FE0000}51.9883}
& 0.2936
& {\color[HTML]{FE0000}21.7120} \\

CPIFuse
& 173.8998
& 0.8740
& {\color[HTML]{34CDF9}110.8920} \\

LFDT
& 1936.6296
& 17.8980
& 692.8200 \\

PAPIF
& 287.4564
& {\color[HTML]{34CDF9}0.2609}
& 323.3230 \\

PIPFNet
& {\color[HTML]{34CDF9}76.3328}
& {\color[HTML]{FE0000}0.0343}
& 288.6660 \\

TIPFNet
& 979.7782
& 4.1509
& 485.6140 \\

DT-F
& 737.8300
& 0.8344
& 4951.9010 \\

\bottomrule
\end{tabular}
}
}
\end{table}
\section{Conclusion}
\label{sec:conclusion}

In this paper, we presented LG-PF, a lightweight confidence-guided framework for polarization image fusion. By treating $S_0$ as the photometric and structural anchor and $DoLP$ as a spatially varying source of complementary information, LG-PF formulates fusion as a confidence-guided residual transfer process. PCP estimates the spatial reliability of polarization responses, MMF regulates their transfer across multiple scales, and LCH performs bounded local correction to stabilize photometric and structural transitions. The same confidence prior is further incorporated into the optimization objectives to preserve stable $S_0$ structures while selectively retaining reliable polarization-sensitive details.

We also constructed MSP, a polarization fusion dataset containing 1000 pixel-aligned $S_0$--$DoLP$ image pairs from 17 indoor and outdoor scene categories. LG-PF achieves the best results across all six evaluated metrics on the MSP test set, while evaluations on fixed subsets of PIF and GAND indicate promising cross-dataset transferability without fine-tuning. Ablation and sensitivity studies further validate the contributions and stability of the proposed design. With only $0.2936$ M parameters, $51.9883$ G FLOPs, and an inference time of $21.712$ ms, LG-PF provides a favorable balance between fusion quality and computational efficiency. Future work will extend MSP to broader materials and acquisition conditions and investigate real-time polarization fusion on resource-constrained imaging devices.

\printcredits

\bibliographystyle{cas-model2-names}

\bibliography{references}


\end{document}